\documentclass[letterpaper, 10 pt, journal, twoside]{IEEEtran}

\makeatletter
\def\@IEEEfigurecaptionsepspace{\vskip 0pt plus 0pt minus 0pt\relax}
\makeatother

\usepackage{booktabs}
\usepackage{cite}
\usepackage{amsmath,amssymb,amsfonts}
\usepackage{algorithm}
\usepackage{algorithmic}
\usepackage{graphicx}
\usepackage{textcomp}
\usepackage{xcolor}
\usepackage{multirow}
\usepackage{adjustbox}
\usepackage{makecell}
\usepackage[hidelinks]{hyperref}

\newcommand{\setvalue}[2]{\expandafter\def\csname myvalue#1\endcsname{#2}}

\newcommand{\getvalue}[1]{\csname myvalue#1\endcsname}

\setvalue{apeTrlOurs}{$\mathbf{0.04(0.02)}$}
\setvalue{apeTrlHinduja}{$0.18(0.13)$}
\setvalue{apeTrlZhang}{$0.57(0.13)$}
\setvalue{apeTrlSwitch}{$0.78(0.18)$}
\setvalue{apeRotOurs}{$\mathbf{0.48(0.38)}$}
\setvalue{apeRotHinduja}{$1.28(0.63)$}
\setvalue{apeRotZhang}{$0.60(0.52)$}
\setvalue{apeRotSwitch}{$0.79(0.57)$}
\setvalue{rpeTrlOurs}{$0.01(0.01)$}
\setvalue{rpeTrlHinduja}{$0.01(0.01)$}
\setvalue{rpeTrlZhang}{$0.01(0.01)$}
\setvalue{rpeTrlSwitch}{$0.01(0.01)$}
\setvalue{rpeRotOurs}{$\mathbf{0.17(0.20)}$}
\setvalue{rpeRotHinduja}{$0.23(0.31)$}
\setvalue{rpeRotZhang}{$0.17(0.21)$}
\setvalue{rpeRotSwitch}{$0.17(0.21)$}
\setvalue{apeTrlOurs2}{$\mathbf{0.02(0.01)}$}
\setvalue{apeTrlHinduja2}{$\mathbf{0.02(0.01)}$}
\setvalue{apeTrlZhang2}{$0.53(0.12)$}
\setvalue{apeTrlSwitch2}{$0.75(0.17)$}
\setvalue{apeRotOurs2}{$\mathbf{0.38(0.18)}$}
\setvalue{apeRotHinduja2}{$\mathbf{0.38(0.18)}$}
\setvalue{apeRotZhang2}{$0.70(0.17)$}
\setvalue{apeRotSwitch2}{$0.84(0.16)$}
\setvalue{rpeTrlOurs2}{$\mathbf{0.01(0.01)}$}
\setvalue{rpeTrlHinduja2}{$\mathbf{0.01(0.01)}$}
\setvalue{rpeTrlZhang2}{$0.01(0.03)$}
\setvalue{rpeTrlSwitch2}{$0.01(0.04)$}
\setvalue{rpeRotOurs2}{$\mathbf{0.12(0.19)}$}
\setvalue{rpeRotHinduja2}{$\mathbf{0.12(0.19)}$}
\setvalue{rpeRotZhang2}{$0.12(0.19)$}
\setvalue{rpeRotSwitch2}{$0.12(0.19)$}

\def\hlfon{0} 
\def\hlf#1{%
  \ifnum\hlfon=1
    \textcolor{blue}{#1}%
  \else
    #1%
  \fi
}

\usepackage{enumitem}

\def\BibTeX{{\rm B\kern-.05em{\sc i\kern-.025em b}\kern-.08em
    T\kern-.1667em\lower.7ex\hbox{E}\kern-.125emX}}

\newcommand{\bmat}[1]{\begin{bmatrix}#1\end{bmatrix}}
\newcommand{\norm}[1]{{\lVert}#1{\rVert}}
\newcommand{\diag}{\text{diag}}

\newif\ifshowcopyright
\showcopyrighttrue 
\ifshowcopyright
\usepackage{fancyhdr}
\fi

\begin{document}

\title{Probabilistic Degeneracy Detection for Point-to-Plane Error Minimization}
\author{Johan Hatleskog and Kostas Alexis\vspace{-1em}
\thanks{\hlf{Manuscript received: May 16th, 2024; Revised: September 1st, 2024; Accepted: October 4th, 2024.}}%
\thanks{\hlf{This paper was recommended for publication by Editor Lucia Pallottino upon evaluation of the Associate Editor and Reviewers' comments.
This work was supported by the Research Council of Norway under awards NO-338694 and NO-310255. (\emph{Corresponding author: Johan Hatleskog.})}}%
\thanks{\hlf{Johan Hatleskog and Kostas Alexis are with the Department of Engineering Cybernetics, Norwegian University of Science and Technology, Trondheim, Norway {\tt\footnotesize johan.hatleskog@gmail.com}}}%
\thanks{\hlf{Johan Hatleskog is also with Cognite AS, Lysaker, Norway {\tt\footnotesize johan.hatleskog@cognite.com}}}%
\thanks{\hlf{Digital Object Identifier (DOI): see top of this page.}}
}

\markboth{\hlf{IEEE Robotics and Automation Letters. Preprint Version. Accepted October, 2024}}
{\hlf{Hatleskog \MakeLowercase{\textit{et al.}}: Probabilistic Degeneracy Detection for Point-to-Plane Error Minimization}}

\maketitle

\ifshowcopyright
\thispagestyle{fancy}
\fi

\begin{abstract}
Degeneracies arising from uninformative geometry are known to deteriorate LiDAR-based localization and mapping. This work introduces a new probabilistic method to detect and mitigate the effect of degeneracies in point-to-plane error minimization. The noise on the Hessian of the point-to-plane optimization problem is characterized by the noise on points and surface normals used in its construction. We exploit this characterization to quantify the probability of a direction being degenerate. The degeneracy-detection procedure is used in a new real-time degeneracy-aware iterative closest point algorithm for LiDAR registration, in which we smoothly attenuate updates in degenerate directions. The method's parameters are selected based on the noise characteristics provided in the LiDAR's datasheet. We validate the approach in four real-world experiments, demonstrating that it outperforms state-of-the-art methods at detecting and mitigating the adverse effects of degeneracies. For the benefit of the community, we release the code for the method at: \href{https://github.com/ntnu-arl/drpm}{github.com/ntnu-arl/drpm}.
\end{abstract}
\section{Introduction}%

\IEEEPARstart{L}{iDAR} is the main sensor modality in a wide range of simultaneous localization and mapping (SLAM) frameworks \cite{zhang2014loam,shan2020lio, liu2021balm,khattak2020complementary,zhang2018laser}. Typically, the incoming LiDAR point cloud is registered to a reference point cloud by minimizing geometric cost functions with variations of the iterative closest point (ICP) algorithm~\cite{besl1992method}. This optimization is prone to failure in environments with uninformative geometry, such as self-similar tunnels and cylindrical tanks \cite{gelfand2003geometrically}. Then, the geometry cannot constrain the full 6-DoF pose, and the optimization is rendered degenerate in unconstrained directions. Due to noise-induced spurious information, the optimization may still yield over-confident and erroneous estimates for unconstrained directions. This necessitates explicit handling of degeneracies for reliable state estimation. 

\subsection{Related Work}

Multiple approaches address this challenge. The works in \cite{nobili2018predicting,nubert2022learning} propose learning-based methods for degeneracy detection. While effective once trained, the reliance on labeled ground truth data limits their generalizability. Additionally, computational demands preclude their use in real-time state estimation on resource-constrained systems.

Degeneracy detection is closely related to covariance estimation for ICP-based registration. Analytic  \cite{censi2007accurate, brossard2020new} and data-driven \cite{landry2019cello,de2022deep} methods have been proposed. However, the need for dedicated degeneracy detection methods stems in part from the difficulty in estimating ICP covariance.

\begin{figure}[t]
    \centering
    \includegraphics[width=1\linewidth]{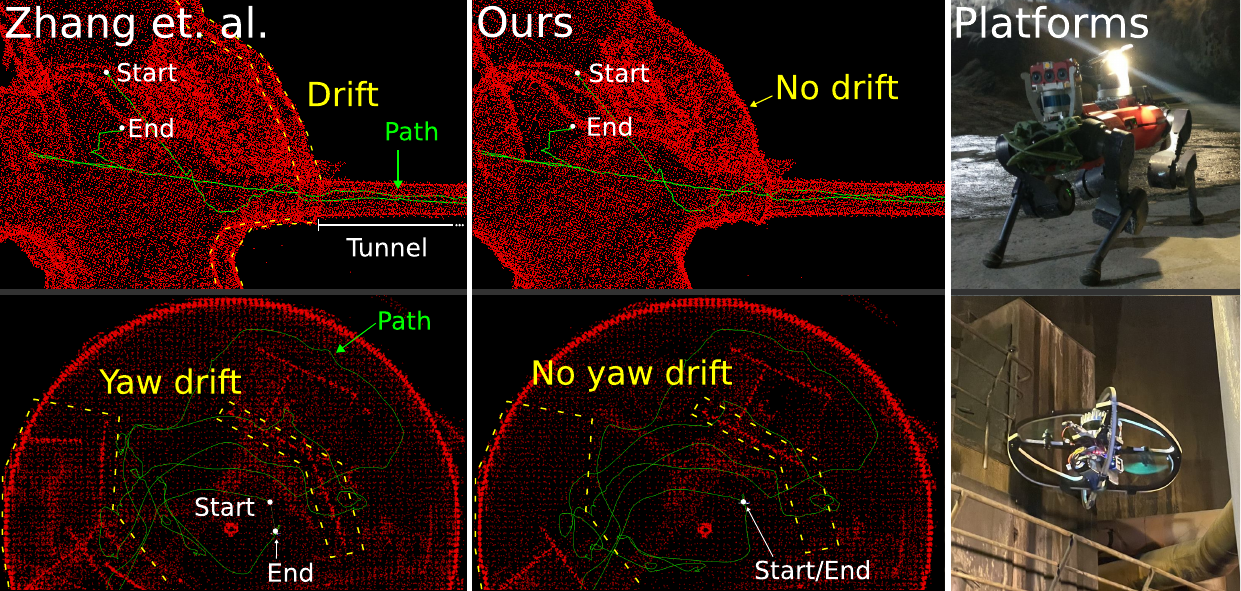}
    \caption{The comparison of our proposed approach (middle column) to the state-of-the-art method from \cite{zhang2016degeneracy} (left column). Our method avoids drift by successfully detecting and accounting for degeneracies. \textbf{Top row:} From the Seem\"uhle underground mine using the legged ANYmal platform (top right). \textbf{Bottom row:} From a cylindrical tank using a flying platform (bottom right).}\label{fig:teaser-figure}
    \vspace{-7mm}
\end{figure}

The seminal work \cite{zhang2016degeneracy} analyzes degeneracies geometrically. A threshold on the eigenvalues of the Hessian of the geometric cost functions is used to detect degeneracy. The authors also introduce the solution remapping technique to only update estimates in non-degenerate directions. \cite{zhang2016degeneracy} is the established state-of-the-art, and is used by SLAM-frameworks such as \cite{khattak2020complementary,zhang2018laser}. \cite{hinduja2019degeneracy} builds on \cite{zhang2016degeneracy} for degeneracy-aware point-to-plane ICP, and incorporates the ICP result in factor graph optimization by proposing a degeneracy-aware factor that only influences non-degenerate directions. Also based on considering eigenvalues, the SLAM-frameworks in \cite{tagliabue2021lion,ebadi2021dare} detect degeneracy by using a threshold on the condition number of the Hessian. However, the methods in \cite{zhang2016degeneracy,hinduja2019degeneracy,tagliabue2021lion,ebadi2021dare} require careful fine-tuning tailored to the robot, sensor, environment and particulars of the localization pipeline they are used in \cite{ebadi2023present}. The eigenvalues of the Hessian depend on the number of measurements, weights, the scale of the environment and sensor noise. Consequently, tuning parameters cannot be expected to generalize across heterogeneous scenarios.\cite{hinduja2019degeneracy,lee2024switch} propose dynamic parameters for \cite{zhang2016degeneracy}, but offer limited generalizability as only a subset of the sources of variation in eigenvalues is accounted for. The localizability metric of \cite{tuna2023x} accounts for scale. However, the method considers the rotational and translational blocks of the Hessian separately and may fail when the full Hessian is degenerate while the individual blocks are not. 

\subsection{Proposed Contribution}
We propose to detect degeneracies by considering a probabilistic characterization of the noise entering the Hessian. The noise characterization is used for soft degeneracy classification in which the probability of a direction being effectively non-degenerate is quantified as the probability of the signal being at least an order of magnitude higher than the noise. The soft classification is subsequently used for probabilistic degeneracy-aware point-to-plane ICP where updates in degenerate directions are attenuated. 

Owing to the first-principles probabilistic formulation, the method explicitly accounts for environment scale, weights and measurement count. The simultaneous generalization across these factors improves on \cite{zhang2016degeneracy,hinduja2019degeneracy,tagliabue2021lion,ebadi2021dare}. The baseline noise level is set via user-provided noise models for the points and normals used in the point-to-plane cost functions. Consequently, the method explicitly accounts for measurement noise and inaccuracies in the surface normals. We select noise models based on the LiDAR's datasheet. 

Similar to \cite{zhang2016degeneracy}, the method is a real-time compatible plug-in step and can be incorporated in common non-linear least squares point-to-plane optimizers while retaining the overall worst-case computational complexity. We incorporate our approach in \cite{khattak2020complementary} --which builds upon LOAM~\cite{zhang2014loam} for LiDAR registration-- for real-time state estimation.

The approach is validated on four real-world datasets collected by legged and aerial robots in challenging environments. We compare against the state-of-the-art methods \cite{zhang2016degeneracy,hinduja2019degeneracy} and the recent method \cite{lee2024switch}. Fig. \ref{fig:teaser-figure} shows indicative results. The supplementary video\footnote{[Online]. Available: \href{https://youtu.be/bKnHs\_wwnXs}{\hlf{https://youtu.be/bKnHs\_wwnXs}}} summarizes the approach and experiments.

Overall, our main contributions are:
\begin{enumerate}[leftmargin=*]
    \item A new probabilistic and generalizable degeneracy detection method for point-to-plane cost functions.
    \item A probabilistic degeneracy-aware point-to-plane ICP algorithm for real-time LiDAR registration that reduces the influence of noise in effectively degenerate directions.
    \item Experiments validating the approach on four real-world datasets, outperforming the methods of \cite{zhang2016degeneracy,hinduja2019degeneracy,lee2024switch}.
\end{enumerate}
As our degeneracy detection method requires the estimation of covariance for normals, we additionally include a practical method to do so.

We continue with preliminaries and the problem statement in section \ref{sec:preliminaries-and-problem-statements}, discuss our approach in section \ref{sec:approach}, present experiments in section \ref{sec:experiments}, and conclude in section \ref{sec:conclusion}.

\section{Preliminaries \& Problem Statement}\label{sec:preliminaries-and-problem-statements}

This section defines the point-to-plane optimization problem and noise models. We subsequently illustrate how noise induces spurious information in degenerate directions, motivating the problem statement.

\subsection{Notation}
$\norm{x}_{2}$ is the Euclidean norm of the vector $x$. $\norm{x}_{\mathbf{S}}$ is the Mahalanobis distance, with $\mathbf{S}>0$. $[a]_{\times}$ is the skew-symmetric matrix representation of vector $a \in \mathbb{R}^3$ such that $[a]_{\times} b = a \times b$ is the cross-product of $a$ and $b \in \mathbb{R}^3$. $\text{SO}(3)$ denotes the special orthogonal group of order 3. $\text{Exp}(x)$ is the capitalized exponential map as defined in \cite{sola2018micro}. Matrices are denoted by bold uppercase letters, e.g. $\mathbf{H}$. $\mathbf{H}^{\dagger}$ is the Moore-Penrose inverse of $\mathbf{H}$. $\mathbf{I}$ is an identity matrix. $\mathcal{N}(\mu, \mathbf{\Sigma})$ denotes a multi-variate normal distribution with mean $\mu$ and covariance $\mathbf{\Sigma}$.

\subsection{Noise-free Point-to-Plane Optimization}\label{sec:p2p-noise-free}

Consider a set of noise-free points $p_i^L \in \mathbb{R}^3$ expressed in the LiDAR frame $L$ and associated noise-free planes $(n_i, d_i) \in \mathbb{R}^3 \times \mathbb{R}$ for $i = 1, \dots, N$ expressed in the world frame. $n_i$ is the unit length plane normal and $d_i$ is the signed orthogonal distance to the plane from the origin. $\mathbf{T} = (\mathbf{R},t)$ is the estimated pose of the LiDAR expressed in the world frame with rotation $\mathbf{R} \in \text{SO}(3)$ and translation $t \in \mathbb{R}^3$. Correspondingly, $p_i = \mathbf{R} p_i^L + t$ is the estimated position of $p_i^L$ in the world frame. The $i$'th point-to-plane error is 
\begin{align}
    r_i(\mathbf{R},t) = n_i^T (\mathbf{R} p_i^L +t) - d_i.
\end{align}
$\mathbf{T}$ is found through the point-to-plane error minimization
\begin{align}\label{eq:point-to-plane-min}
    \min_{ \mathbf{R}, t} \sum_{i=1}^N \rho(\norm{w_{r_i} r_i(\mathbf{R},t)}_{2}),
\end{align}
where $\rho(u)$ is an M-estimator class cost function such as the standard $L^2$ norm, $\rho(u)= u^2/2$, or sub-quadratic robust variants such as Geman-McClure, $\rho(u) = \frac{1}{2} \frac{u^2}{1+u^2}$. See \cite{mactavish2015all} for an overview of popular robust cost functions. The $w_{r_i}$'s are non-negative scalar weights. Following \cite{pomerleau2015review}, we linearize \eqref{eq:point-to-plane-min} about an initial guess of $\mathbf{T}$ and formulate the optimization as the least squares problem
\begin{align}
    \min_x \sum_i \norm{ \mathbf{J}_i x - b_i}_{2}^2, \label{eq:point-to-plane-linearized}
\end{align}
with $\mathbf{J}_i = w_i \bmat{ (p_i \times n_i)^T & n_i^T }$ and ${b_i = - w_i (n_i^T p_i - d_i)}$. The optimization variable $x \in \mathbb{R}^6$ is the twist $x = [\delta r^T, \delta t^T]^T$ with $\delta r$ and $\delta t$ representing the rotational and translational perturbation, respectively. $w_i$ is defined as $w_i = w_{r_i} w_{\rho_i}$, where $w_{\rho_i}$ is the attenuation of the $i$'th error due to the choice of cost function. With $u_i = \norm{w_{r_i} r_i}_2$, then $w_{\rho_i}^2 = \rho'(u_i) / u_i$. E.g. $w_{\rho_i}^2=1$ for the $L^2$ norm and $w_{\rho_i}^2= (1+u_i^2)^{-2}$ for the Geman-McClure cost function \cite{mactavish2015all}. An optimal perturbation $x^\star$ is found by solving
\begin{align}
    \left (\sum_i \mathbf{J}_i^T \mathbf{J}_i \right )x = \sum_i \mathbf{J}_i^T b_i.\label{eq:linearized-problem}
\end{align}
$\mathbf{T}$ is updated through $\mathbf{T} \leftarrow \text{Exp}(x^\star) \mathbf{T}$. The problem \eqref{eq:point-to-plane-min} is solved using standard ICP, whereby the process of associating points and planes, linearizing about the current estimate of $\mathbf{T}$, solving for the optimal perturbation and updating the estimate is repeated until a termination criterion is met. 

Use $\mathbf{H}$ to denote the Hessian of the optimization problem,
\begin{align}\label{eq:hessian-noise-free}
    \mathbf{H} = \sum_i \mathbf{J}_i^T \mathbf{J}_i = \sum_i w_i^2 \bmat{ [p_i]_{\times} \\ \mathbf{I} } n_i n_i^T \bmat{ [p_i]_{\times}^T & \mathbf{I}^T }.
\end{align}
Degeneracy entails that the cost functions in \eqref{eq:point-to-plane-min} provide no information in at least one direction of the state space. Accordingly, there exists at least one direction $u$ such that $u^T \mathbf{H} u = 0$. In this case, we define the noise-free perturbation $x^\star$ as the least-norm solution to \eqref{eq:linearized-problem},
\begin{align}
    x^\star = \left (\sum_i \mathbf{J}_i^T \mathbf{J}_i \right )^\dagger \sum_i \mathbf{J}_i^T b_i.\label{eq:noise-free-moore-penrose}
\end{align}
Then $u^T x^\star = 0$ for any direction $u$ such that $u^T \mathbf{H} u = 0$ and no spurious information is present. In practice, spurious information enters the estimate of $x^\star$ due to noise. 
 
\subsection{Noise Models}

We assume that the points $\hat{p}_i$ are affected by additive Gaussian noise
\begin{align}\label{eq:noise- model-point}
    \hat{p}_i = p_i + \epsilon_i, \quad \epsilon_i \sim \mathcal{N}(0, \mathbf{\Sigma}_{p_i}).
\end{align}
The normals are assumed to be influenced by a small rotational perturbation $\Bar{\mathbf{R}}_i$  such that, using the small angle approximation, 
\begin{align}\label{eq:noise- model-normal}
    \hat{n}_i = \bar{\mathbf{R}}_i^T n_i \approx [\mathbf{I} - [\eta_i]_{ \times }] n_i = n_i + [n_i]_{ \times} \eta_i
\end{align}
with $\eta_i \sim \mathcal{N}(0, \mathbf{\Sigma}_{n_i})$. All $\eta_i$'s and $\epsilon_i$'s are assumed to be mutually independent.

To ensure that the Gaussian error assumption is sufficiently satisfied for points, normals and point-to-plane errors, robustness measures such as outlier rejection and robust cost function should be used as necessary. See \cite{pomerleau2015review} for a summary of outlier filtering methods for ICP pipelines. 

\subsection{On Noise-Induced Spurious Information}\label{sec:on-noise-induced-spurious-information}

To motivate our approach, we exemplify how noise on normals induces spurious information in the Hessian. Assume that $\mathbf{\Sigma}_{n_i} = \sigma_n^2 \mathbf{I}$ and $\mathbf{\Sigma}_{p_i} \approx \mathbf{0}$.  Consider the normal equation \eqref{eq:linearized-problem} with $\hat{n}_i$'s in place of $n_i$'s, 
\begin{align}
    \sum_i \mathbf{F}_i \hat{n}_i \hat{n}_i^T \mathbf{F}_i^T \hat{x} = \sum_i w_i \mathbf{F}_i \hat{n}_i \hat{n}_i^T (p_i - q_i), \label{eq:normal-equation-with-noise-on-normals}   
\end{align}
with $\mathbf{F}_i^T = w_i \bmat{ [p_i]_{\times}^T & \mathbf{I} }$ such that $\mathbf{J}_i = n_i^T \mathbf{F}_i^T$. $q_i$ is an arbitrary point on plane $i$. Applying the conditional expectation $E[\cdot | \hat{x}]$ on both sides of \eqref{eq:normal-equation-with-noise-on-normals} yields
\begin{equation}\label{eq:expected-normal-eq}
\begin{split}
    \sum_i \mathbf{F}_i[ &n_i n_i^T  + \sigma_n^2 (\mathbf{I} - n_i n_i^T)] \mathbf{F}_i^T \hat{x} = \\ &\sum_i w_i \mathbf{F}_i (n_i n_i^T + \sigma_n^2 (\mathbf{I} - n_i n_i^T)) (p_i - q_i).
\end{split}
\end{equation}
We additionally define $\mathbf{H}_N =\sum_i \sigma_n^2 \mathbf{F}_i (\mathbf{I}-n_i n_i^T) \mathbf{F}_i^T$ and ${x_N=\mathbf{H}_N^{\dagger} \sum_i \sigma_n^2 w_i \mathbf{F}_i (\mathbf{I}-n_i n_i^T) (p_i - q_i)}$. Then, \eqref{eq:expected-normal-eq} can be stated as
\begin{align}
    (\mathbf{H} + \mathbf{H}_N) \hat{x} = \mathbf{H} x^\star + \mathbf{H}_N x_N.\label{eq:information-weighting-example}
\end{align}
We, therefore, expect the estimated perturbation $\hat{x}^\star$ to be a weighted average of the noise-free perturbation $x^\star$ and bias from $x_N$. If $\mathbf{H} \gg \mathbf{H}_N$, then $\hat{x}^{\star} \approx x^\star$ as desired. If, on the other hand, uninformative geometry renders the noise-free Hessian degenerate in some direction $u$ such that  $u^T \mathbf{H} u = 0$, then $u^T \hat{x}^\star$ is solely noise. Similarly, we expect the estimate in direction $u$, $u^T \hat{x}^\star$, to be unreliable if the spurious information in $u^T \mathbf{H}_N u$ dominates the true information in $u^T \mathbf{H} u$. By detecting and damping directions where the noise in $\mathbf{H}_N$ dominates the signal in $\mathbf{H}$, one can reduce the noise-induced bias from $x_N$. This is the approach we pursue.

\subsection{Problem Statement}

Motivated by the preceding discussion, this work addresses the following problems:
\begin{enumerate}[leftmargin=*]
    \item Given a noisy point-to-plane Hessian $\hat{\mathbf{H}}$ and noise models for the points and normals used in its construction, estimate the noise affecting $\hat{\mathbf{H}}$ and quantify the probability that a direction $u$ of $\hat{\mathbf{H}}$ is degenerate. 
    \item Improve point-to-plane optimization by using the degeneracy probabilities to reduce the influence of spurious noise in directions deemed degenerate.
\end{enumerate}

\section{Approach}\label{sec:approach}

\subsection{Probabilistic Degeneracy Detection}\label{sec:prob-degen-detection}
The goal of probabilistic degeneracy detection is to quantify the probability of the Hessian being degenerate in an arbitrary direction $u \in \mathbb{R}^6$. To that end, we estimate the noise entering the Hessian and assess a signal-to-noise ratio.  

The noisy Hessian $\hat{\mathbf{H}}$ is a sum of outer products of vectors
\begin{align}
    \hat{\mathbf{H}} &= \sum_i \hat{v}_i \hat{v}_i^T, \quad \hat{v}_i = w_i \bmat{ [\hat{p}_i]_{ \times } \\ \mathbf{I} } \hat{n}_i. \label{eq:noisy-hessian-and-vector-v}
\end{align}
The noise characteristics of the outer products are determined by the noise characteristics of the vectors they comprise. Using the noise models \eqref{eq:noise- model-point}-\eqref{eq:noise- model-normal} with $\hat{v}_i$ in \eqref{eq:noisy-hessian-and-vector-v} yields
\begin{equation}\label{eq:vector-v-noise-model}
\begin{split}
        \hat{v}_i = &v_i + \underbrace{ {w_i} \bmat{ - [n_i]_{ \times } & [p_i]_{ \times } [n_i]_{\times} \\ 0 & [n_i]_{\times}} }_{:=\mathbf{B}_i} \bmat{ \epsilon_i \\ \eta_i } 
    \\ &+ {w_i} [\epsilon_i]_{\times} [n_i]_{\times} \eta_i, \quad v_i = {w_i} \bmat{ [p_i]_{ \times } \\ \mathbf{I} } n_i.
\end{split}
\end{equation}
The expectation $E[\hat{v}_i]$ and the covariance $\mathbf{\Sigma}_i = \text{cov}(\hat{v}_i)$ are
\begin{align}
    E[\hat{v}_i] = v_i,\quad \mathbf{\Sigma}_i = \mathbf{B}_i \bmat{ E[\epsilon_i \epsilon_i^T] & 0 \\ 0 & E[\eta_i \eta_i^T]  } \mathbf{B}_i^T.
\end{align}
In turn, the expectation of the noisy Hessian in \eqref{eq:noisy-hessian-and-vector-v} is
\begin{equation}\label{eq:expectation-of-info-matrix}
\begin{split}
    E[\hat{\mathbf{H}}] &= \sum_i E[\hat{v}_i \hat{v}_i^T] = \sum_i v_i v_i^T + \sum_i \text{cov}(\hat{v}_i) \\
    &= \mathbf{H} + \mathbf{\Sigma}, \quad \mathbf{\Sigma} := \sum_i \text{cov}(\hat{v}_i).
\end{split}
\end{equation}
That is, in expectation $\mathbf{H}$ is perturbed by $\mathbf{\Sigma}$. We compute the variance of $\hat{\mathbf{H}}$ in direction $u$, $\norm{u}_2=1$, as
\begin{align}\label{eq:cov-of-sum-of-outer-product}
        \sigma_u^2 &= \text{cov}(u^T \hat{\mathbf{H}} u) = \sum_i \text{cov}(u^T \hat{v}_i \hat{v}_i^T u).
\end{align}
Since $\text{cov}(u^T \hat{v}_i \hat{v}_i^T u) = \text{cov}( \hat{v}_i^T u u^T \hat{v}_i)$, we use the expression for the variance of quadratic forms in \cite{petersen2008matrix} to find
\begin{align}\label{eq:cov-of-outer-product}
    \text{cov}(u^T \hat{v}_i \hat{v}_i^T u) &= 2(u^T \mathbf{\Sigma}_i u)^2 + 4 (u^T \mathbf{\Sigma}_i u) (u^T v_i)^2.
\end{align}
Note that if the $\hat{v}_i$'s were scalar with unit variance, then \eqref{eq:expectation-of-info-matrix} and \eqref{eq:cov-of-sum-of-outer-product} would correspond to the mean and variance of a non-central chi-squared distribution \cite{krishnamoorthy2006handbook}, respectively.  

The expectation and variance in \eqref{eq:expectation-of-info-matrix}-\eqref{eq:cov-of-sum-of-outer-product} are used to quantify the likelihood of a direction being degenerate. As the true points and normals are unavailable to us in practice, we use their noisy counterparts in estimates of $\mathbf{\Sigma}_i$, $\mathbf{\Sigma}$ and $\sigma_u^2$. These estimates are denoted by $\hat{\mathbf{\Sigma}}_i$, $\hat{\mathbf{\Sigma}}$ and $\hat{\sigma}_u^2$, respectively. Let $\xi_u$ denote the noise in direction $u$, and assume that $\xi_u \sim \mathcal{N}(\hat{\mu}_u, \hat{\sigma}_u^2)$ with
\begin{align}
    \hat{\mu}_u &= u^T \hat{\mathbf{\Sigma}} u, \\ 
    \hat{\sigma}_u^2 &= \sum_i [2(u^T \hat{\mathbf{\Sigma}}_i u)^2 + 4 (u^T \hat{\mathbf{\Sigma}}_i u) (u^T \hat{v}_i)^2].
\end{align}
Moreover, let $a_{u} = u^T \mathbf{H} u$ and $\hat{a}_{u} = u^T \hat{\mathbf{H}} u$. By definition, we have $\hat{a}_u = a_u + \xi_u$. $a_{u} / \xi_u $ is a signal-to-noise ratio. As discussed in section \ref{sec:preliminaries-and-problem-statements}, the signal $a_u$ should dominate the noise $\xi_u$ for direction $u$ to be reliably estimated. Thus, we estimate the probability that the signal is $s$ times larger than noise in direction $u$ as
\begin{align}
    p_u = P(a_{u} \geq s \xi_u ) = P(\hat{a}_u \geq (s+1) \xi_u ). \label{eq:degen-probability}
\end{align}
To set $s$, note that the relative error $e_u^{\text{rel}} = (\hat{a}_u - a_u)/a_u = \xi_u / a_u$ is bounded by $s$ in the sense that if $a_u \geq s \xi_u$ then $e_u^{\text{rel}} \leq s^{-1}$. We set $s=10$ to target a relative error of at most $10 \%$. Generally, increasing $s$ makes the degeneracy detection more conservative. $u$ is considered degenerate if $p_u \approx 0$ and non-degenerate if $p_u \approx 1$.

Observe from \eqref{eq:noisy-hessian-and-vector-v}-\eqref{eq:expectation-of-info-matrix} that the noise entering the Hessian depends on the measurement count $N$, weights $\{w_i\}_{i=1, \dots, N}$, noise from points and normals in \eqref{eq:noise- model-point}-\eqref{eq:noise- model-normal} and, through the cross-product $[\hat{p}_i]_{ \times } \hat{n}_i$, the environment scale. Unlike previous approaches \cite{zhang2016degeneracy,hinduja2019degeneracy,tagliabue2021lion,ebadi2021dare,tuna2023x}, the probability $p_u$ in \eqref{eq:degen-probability} simultaneously accounts for all these factors.

\subsection{Probabilistic Degeneracy-Aware ICP}

The goal of probabilistic degeneracy-aware ICP is to improve point-to-plane ICP \cite{low2004linear} by reducing the influence of spurious information in degenerate directions. We limit the discussion to the specific changes we make to standard point-to-plane ICP and refer the reader to \cite{pomerleau2015review} for an in-depth treatment of the full ICP point cloud registration pipeline. Similar to \cite{zhang2016degeneracy, hinduja2019degeneracy}, we adapt the update step of the ICP algorithm.

As a motivation, we revisit the noise-free case discussed in section \ref{sec:p2p-noise-free} and consider specifically the Moore-Penrose inverse of $\mathbf{H}$. Use the diagonalization $\mathbf{H} = \mathbf{U} \mathbf{\Lambda} \mathbf{U}^T$, where $\mathbf{U} = [u_1, \cdots, u_6]$ is an orthonormal matrix such that $\mathbf{U}^T \mathbf{U} = \mathbf{I}$ and $\mathbf{\Lambda}=\text{diag}(\lambda_1, \cdots, \lambda_6)$. The Moore-Penrose inverse of $\mathbf{H}$ is ${\mathbf{H}}^\dagger = \mathbf{U} \mathbf{\Lambda}^\dagger \mathbf{U}^T$ where $\mathbf{\Lambda}^\dagger = \text{diag}(\lambda_1^\dagger, \cdots, \lambda_6^\dagger)$ and
\begin{align}\label{eq:psinv-dagger-plus}
    \lambda_k^\dagger = \begin{cases} 
      \frac{1}{\lambda_k}, &  \lambda_k > 0 \\
    0, & \lambda_k = 0
    \end{cases}.
\end{align}
$\lambda_{k}$ and $\lambda_k^\dagger$ are zero if direction $u_k$ is degenerate. Consequently, using the minimum-norm perturbation $x^\star$ in \eqref{eq:noise-free-moore-penrose} ensures that degenerate directions are left unchanged in the sense that $u_k^T x^{\star} = 0$ if $\lambda_k = 0$.

To account for noise, we define a modified pseudo-inverse. The diagonalization of the noisy Hessian is $\hat{\mathbf{H}} = \hat{\mathbf{U}} \hat{\mathbf{\Lambda}} \hat{\mathbf{U}}^T$. The noise makes all eigenvalues of $\hat{\mathbf{H}}$ non-zero with probability 1 when $N \geq 6$, and \eqref{eq:psinv-dagger-plus} is not useful. Instead, define the indicator variable $\mathbb{I}_k \in \{1,0\}$ and let $\mathbb{I}_k=0$ if direction $\hat{u}_k$ is classified as degenerate and $1$ otherwise. We replace \eqref{eq:psinv-dagger-plus} by
\begin{align}\label{eq:psinv-mod-2}
    \hat\lambda_k^\ddagger = \mathbb{I}_k \frac{1}{\hat\lambda_k} + (1 - \mathbb{I}_k) \cdot 0. 
\end{align}
\hspace{1sp}\cite{zhang2016degeneracy,hinduja2019degeneracy} set the degeneracy classification indicator variable $\mathbb{I}_k$ based on a single threshold for all eigenvalues. In particular, $\mathbb{I}_k = 1$ if $\hat{\lambda}_k > \lambda_{ \text{min} }$ for some threshold $\lambda_{ \text{min} }$. This amounts to a truncated singular value decomposition \cite{hansen2010discrete} of $\hat{\mathbf{H}}$. In our case, we use the probabilistic degeneracy detection from section \ref{sec:prob-degen-detection} to determine the probability of $\mathbb{I}_k$ being zero or one for each eigenvector $\hat{u}_k$. We take the expectation of \eqref{eq:psinv-mod-2} with respect to $\mathbb{I}_k$ by using the probability \eqref{eq:degen-probability} to retrieve
\begin{align}
        \hat\lambda_k^+ = E_{\mathbb{I}_k} [\hat\lambda_k^\ddagger] = E_{\mathbb{I}_k} [\mathbb{I}_k \frac{1}{\hat\lambda_k}] = p_{\hat{u}_k} \frac{1}{\hat\lambda_k}. \label{eq:lambda-plus}
\end{align}
\eqref{eq:lambda-plus} accounts for degeneracies in a smooth manner, softening the transition between non-degenerate and degenerate geometries. Let $\mathbf{J}^T = \bmat{\mathbf{J}_1^T \cdots \mathbf{J}_N^T}$ and $b^T = [ b_1 \cdots b_N]^T$ and let $\hat{\mathbf{J}}$ and $\hat{b}$ be their noisy counterparts. Using \eqref{eq:lambda-plus}, we estimate the perturbation as
\begin{align}
    \hat{x}^\star = \hat{\mathbf{U}} \mathbf{P} \hat{\mathbf{\Lambda}}^{-1} \hat{\mathbf{U}}^T \hat{\mathbf{J}}^T \hat{b} \label{eq:pdicp-estimate},
\end{align}
where $\mathbf{P} =\diag (p_{\hat{u}_1}, \dots, p_{\hat{u}_6})$. From a least-squares perspective, this is equivalent to simultaneously reducing the information of the point-to-plane cost functions and adding a zero-prior in directions where $p_{u_k} < 1$. To see this, use the singular value decomposition $\hat{\mathbf{J}}= \hat{\mathbf{V}} \hat{\mathbf{S}} \hat{\mathbf{U}}^T$, $\hat{\mathbf{S}}^T \hat{\mathbf{S}} = \hat{\mathbf{\Lambda}}$, and 
\begin{align}
    \mathbf{W}^{1/2} &= \hat{\mathbf{V}} \text{diag}(\mathbf{P}^{1/2}, \mathbf{I}_{N-6}) \hat{\mathbf{V}}^T.
\end{align}
The minimizer of the cost
\begin{align}
    \norm{ \mathbf{W}^{1/2} \hat{\mathbf{J}} \hat{x} - \mathbf{W}^{1/2} \hat b }_2^2 + \norm{(\mathbf{I}-\mathbf{P})^{1/2} \hat{\mathbf{\Lambda}}^{1/2} \hat{\mathbf{U}}^T \hat x}_2^2, \label{eq:least-squares-interpretation-1}
\end{align}
is found as the solution to
\begin{align}
    \hat{\mathbf{U}} [\mathbf{P} \hat{\mathbf{\Lambda}} + (\mathbf{I} - \mathbf{P}) \hat{\mathbf{\Lambda}}] \hat{\mathbf{U}}^T \hat{x} =  \hat{\mathbf{J}}^T \hat{\mathbf{J}} \hat{x} = \hat{\mathbf{U}} \mathbf{P} \hat{\mathbf{U}}^T \hat{\mathbf{J}}^T  \hat{b} \label{eq:least-squares-interpretation-2}.
    \end{align}
Left multiplying \eqref{eq:least-squares-interpretation-2} by $(\hat{\mathbf{J}}^T \hat{\mathbf{J}})^{-1} = \hat{\mathbf{H}}^{-1}$ yields the same solution as \eqref{eq:pdicp-estimate}. $\mathbf{W}$ inflates the covariance of the point-to-plane errors. The zero-prior in \eqref{eq:least-squares-interpretation-1} is a Tikhonov regularizer \cite{hansen2010discrete} and addresses the ill-posedness of the underlying optimization by penalizing the magnitude of perturbations in directions deemed partially or fully degenerate. 

Based on the least-squares interpretation, we estimate the information matrix (or inverse-covariance matrix) of $\hat{x}^\star$ as
\begin{align}
    \hat{\mathbf{\Lambda}}_{\hat{x}} = (1/\sigma_r^2)\hat{\mathbf{U}} \mathbf{P} \hat{\mathbf{\Lambda}} \hat{\mathbf{U}}^T,\label{eq:information-estimate-for-perturbation}
\end{align}
where $\sigma_r^2$ is an estimate of the variance of the point-to-plane residuals. Note that in \eqref{eq:information-estimate-for-perturbation} the information from the regularizer is disregarded. $u^T\hat{\mathbf{\Lambda}}_{\hat{x}} u = 0$ when $p_u=0$, reflecting the lack of information (and infinite variance) in degenerate directions. 

The approach is summarized in Algorithm \ref{alg:picp-algorithm}. The algorithm achieves the goal of reducing the influence of spurious information from degenerate directions in the sense that if $p_{\hat{u}_k} \approx 0$ then $\hat{u}_k^T \hat{x}^\star \approx 0$. Non-degenerate directions where $p_{\hat{u}_k} \approx 1$ are left unchanged. The added per-update computation time relative to vanilla point-to-plane ICP is $O(N)$, incurred when computing $\hat{\mu}_{\hat{u}_k}$, $\hat{\sigma}_{\hat{u}_k}$ and $p_{\hat{u}_k}$ for $k=1, \dots, 6$. I.e. the overall worst-case computational complexity is the same as for vanilla point-to-plane ICP, making the approach well-suited for real-time operation.

\vspace{-2mm}
\begin{algorithm}
\renewcommand{\algorithmicrequire}{\textbf{Input:}}
\renewcommand{\algorithmicensure}{\textbf{Output:}}
\caption{Update-Step of Degeneracy-Aware ICP}
\label{alg:picp-algorithm}
\begin{algorithmic}[1]
\REQUIRE $\{ \hat{p}_i$, $\hat{n}_i$, $\hat{d}_i$, $w_{r_i}$,  $\mathbf{\Sigma}_{p_i}$, $\mathbf{\Sigma}_{n_i} \}_{i=1, \dots, N} $
\ENSURE $\hat{x}^\star$
\STATE Compute the Hessian $\hat{\mathbf{H}}$ and right hand side $\sum_i \hat{\mathbf{J}}_i^T \hat{b}_i$
\STATE Diagonalize the Hessian $\hat{\mathbf{H}} = \hat{\mathbf{U}} \hat{\mathbf{\Lambda}} \hat{\mathbf{U}}^T$
\STATE Compute $\hat{\mu}_{\hat{u}_k}$, $\hat{\sigma}_{\hat{u}_k}$ and $p_{\hat{u}_k}$ for $k=1, \dots, 6$
\STATE $\hat{x}^\star \leftarrow \hat{\mathbf{U}} \mathbf{P} \hat{\mathbf{\Lambda}}^{-1} \hat{\mathbf{U}}^T \sum_i \hat{\mathbf{J}}_i^T \hat{b}_i$.
\RETURN $\hat{x}^\star$
\end{algorithmic}
\end{algorithm}
\vspace{-3mm}

\subsection{Covariance Estimates for Normals}\label{sec:normal-covariance}

For practical use of our degeneracy-aware ICP, the covariance of the normals must be estimated. We provide a straightforward way to do so. 

Let $\mathcal{Q}_i = \{\hat{q}_{il}\}_{ l= 0, \dots, N_i }$ be the set of $N_i$ points used to estimate normal $i$. Assume that each point $q_{il}$ is influenced by isotropic Gaussian noise, 
\begin{align}
    \hat{q}_{il}= {q}_{il} + \epsilon_{il}, \epsilon_{il} \sim \mathcal{N}(0, \mathbf{I} \sigma_i^2).
\end{align}
All $\epsilon_{il}$'s are assumed to be mutually independent. Let $\hat{\bar{q}}_i$ be the empirical mean of the points in $\mathcal{Q}_i$. The empirical covariance of the points in $\mathcal{Q}_i$ is
\begin{align}
    \hat{\mathbf{C}}_i &= \frac{ 1 }{N_i - 1} \sum_l ( \hat{q}_{il} - \hat{\bar{q}}_i )(\hat{q}_{il} - \hat{\bar{q}}_i)^T. \label{eq:empirical-cov-normal}  
\end{align}
A normal estimate $\hat{n}_i$ is found as the eigenvector corresponding to the smallest eigenvalue of the empirical covariance matrix \eqref{eq:empirical-cov-normal}, as originally proposed in \cite{hoppe1992surface}. We find a covariance estimate for each $\hat{n}_i$ by considering the associated least squares problem. Let $\hat{n}_i = \hat{\mathbf{R}}_i e_z$ be our estimate of the $i$'th normal, for some rotation matrix $\hat{\mathbf{R}}_i \in \text{SO}(3)$ and $e_z = \bmat{0, 0, 1}^T$. $\hat{\mathbf{R}}_i$ is found as an optimizer of the non-linear least squares problem
\begin{equation}\label{eq:min-normal}
\begin{split}
    &\min_{{\mathbf{R}_i} \in \text{SO}(3)} \sum_l \norm{ ( \hat{q}_{il} - \hat{\bar{q}}_i  )^T {\mathbf{R}_i} e_z }_{ \frac{N_i-1}{N_i} \sigma_i^2 }^2 \\
    &= \min_{{\mathbf{R}_i} \in \text{SO}(3)} \frac{1}{ \sigma_i^2 / N_i } \norm{ \hat{\mathbf{C}}_{i}^{1/2} {\mathbf{R}_i} e_z }_2^2.
\end{split}
\end{equation}
An optimizer of \eqref{eq:min-normal} is found by using the diagonalization $\hat{\mathbf{C}}_{i} = \hat{\mathbf{R}}_i \hat{\mathbf{\Lambda}}_i \hat{\mathbf{R}}_i^T$ with the eigenvalues in $\hat{\mathbf{\Lambda}}_i = \diag (\hat{\lambda}_{i1},\hat{\lambda}_{i2},\hat{\lambda}_{i3})$ ordered from large to small and with $\hat{\mathbf{R}} \in \text{SO}(3)$. In particular, $\hat{\mathbf{R}}_i$ is an optimizer of \eqref{eq:min-normal}. We use the Hessian of the linearized optimization problem to find a covariance estimate. Linearizing by using the small-angle approximation $[\mathbf{I} + [\nu_i]_{ \times } ] {\mathbf{R}}_i$ yields the Jacobian
\begin{align}
    \hat{\mathbf{J}}_{n_i} = - \frac{ \hat{\mathbf{C}}_i^{1/2} [\hat{n}_i]_{ \times } }{ \sqrt{\sigma_i^2 / N} }
\end{align}
at $\mathbf{R}_i = \hat{\mathbf{R}}_i$. The Hessian at $\mathbf{R}_i = \hat{\mathbf{R}}_i$ is
\begin{align}
    \hat{\mathbf{J}}_{n_i}^T \hat{\mathbf{J}}_{n_i} = \frac{1}{\sigma_i^2/N} \hat{\mathbf{R}}_i \text{diag}(\hat{\lambda}_{i2}, \hat{\lambda}_{i1}, 0) \hat{\mathbf{R}}_i^T \label{eq:normal-hessian-last}
\end{align}
The pseudo-inverse of \eqref{eq:normal-hessian-last}  yields the covariance estimate
\begin{align}
    \hat{\Bar{\mathbf{\Sigma}}}_{\hat{n}_i} &= (\hat{\mathbf{J}}_{n_i}^T \hat{\mathbf{J}}_{n_i})^{\dagger} = \hat{\mathbf{R}}_i \frac{\sigma_i^2}{N} \text{diag}(\frac{1}{\hat{\lambda}_{i2}}, \frac{1}{\hat{\lambda}_{i1}}, 0) \hat{\mathbf{R}}_i^T. \label{eq:normal-cov-est}
\end{align}
Agreeing with intuition, the covariance of $\hat{n}_i$ increases as the spread of the points in $\mathcal{Q}_i$ in directions orthogonal to $\hat{n}_i$ is reduced. The variance in direction $\hat{n}_i$ is zero as the normal is confined to the unit sphere manifold. 

It can be shown that \eqref{eq:normal-cov-est} is equal to $\hat{\Bar{\mathbf{\Sigma}}}_{\hat{n}_i} = [\hat{n}_i]_{\times} \frac{\sigma_i^2}{N_i} \hat{\mathbf{C}}^{-1}_i  [\hat{n}_i]_{\times}^T$. To adhere with the noise model \eqref{eq:noise- model-normal}, we use the estimate on the form
\begin{align}\label{eq:normal-noise-model}
    \hat{n}_i \approx n_i + [\hat{n}_i]_{\times} \eta_i, \quad \eta_i \sim \mathcal{N}(0, \frac{\sigma_i^2}{N_i} \hat{\mathbf{C}}^{-1}_i ).
\end{align}
The computational complexity of estimating the normal and its covariance is $O(N_i)$, but the added cost of the covariance estimate in \eqref{eq:normal-noise-model}  is only $O(1)$. As such, the covariance estimate is a computationally cheap byproduct of surface normal estimation.

We additionally use the covariance estimate for outlier rejection. From \eqref{eq:normal-cov-est} it is clear that the worst-case variance $\max_{\norm{u}_2=1 }u^T \hat{\Bar{\mathbf{\Sigma}}}_{\hat{n}_i} u$ is given by $\hat{\sigma}_{\hat{n}_i}^2 = (\sigma_i^2/N_i) \hat{\lambda}_{i2}^{-1}$. $\hat{n}_i$ is classified as an outlier if $\hat{\sigma}_{\hat{n}_i}^2 > \sigma_{n_\text{max}}^2$, where $\sigma_{n_\text{max}}^2 > 0$ is a user-supplied threshold. 

\section{Experiments}\label{sec:experiments}

The proposed approach is evaluated on four datasets using legged and aerial robots in challenging environments.

\subsection{Implementation and Parameter Selection}\label{sec:parameter-selection}
The approach in section \ref{sec:approach} is integrated into the CompSLAM framework \cite{khattak2020complementary}, which relies on the LOAM method~\cite{zhang2014loam} as long as LiDAR data are concerned. The LOAM mapping module is adapted as follows: 1) The LOAM ICP update step is replaced by algorithm \ref{alg:picp-algorithm}; 2) we only use planar features; and 3) we estimate the per-normal covariance using the method in section \ref{sec:normal-covariance}. Features with $\hat{\sigma}_{\hat{n}_i}^2 > \sigma_{n_\text{max}}^2 = 0.10^2$ are rejected as outliers. All inputs to algorithm  \ref{alg:picp-algorithm} are expressed in the local LiDAR frame. Adjoint transformations (see \cite{sola2018micro}) are used to transform the resulting local perturbation to the world frame. LOAM odometry is bypassed and the prior for LiDAR mapping is computed using an external odometry source. 

We compare against the state-of-the art methods of \cite{zhang2016degeneracy,hinduja2019degeneracy} and the recent method of \cite{lee2024switch} by implementing them within the same framework. The method of \cite{zhang2016degeneracy} requires access to a sample dataset for tuning, and its parameter is tuned once per platform using the procedure in \cite{zhang2016degeneracy}. We select parameters for our method once per platform based on the LiDAR's datasheet. The datasets in experiments 1, 3 and 4 lack ground truth. Hence, we provide qualitative results for all experiments and quantitative results for experiment 2.

\subsection{Experiment 1 - R\"umlang Construction Site}

\begin{figure}[t]
    \vspace{0.5mm}
    \centering
    \includegraphics[width=1\linewidth]{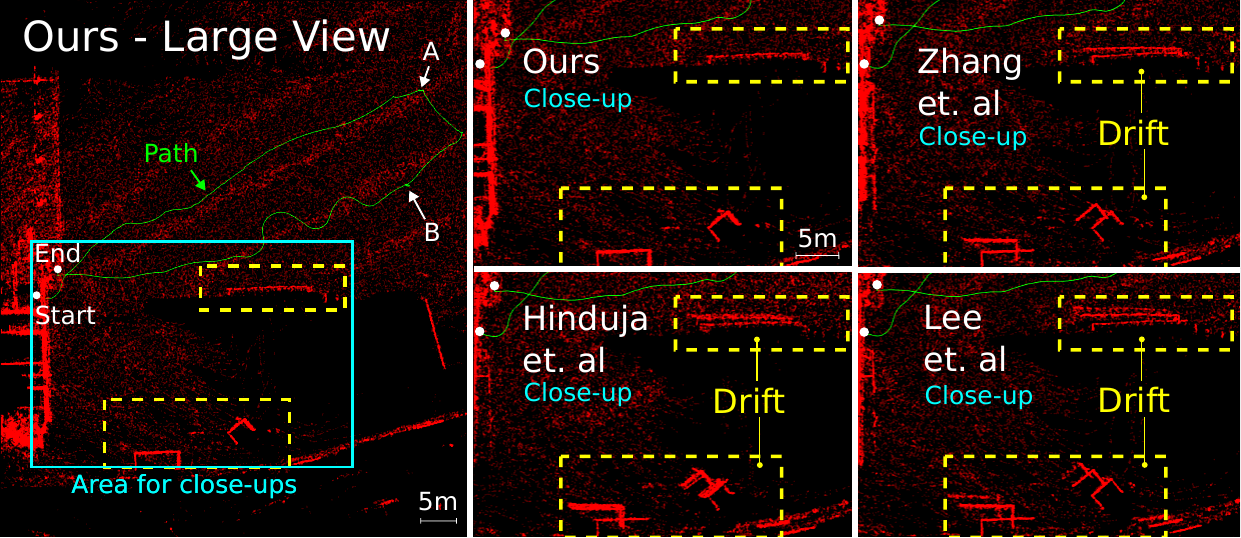}
    \caption{Partial maps of all methods from experiment 1 (R\"umlang). The field-of-view of the Velodyne VLP-16 LiDAR is reduced to $180^\circ$ to study degeneracies. \cite{zhang2016degeneracy,hinduja2019degeneracy,lee2024switch} exhibit degeneracy-induced drift, materializing as erroneously repeated structures in the close-ups.}
    \label{fig:rumlang-result}
    \vspace{-5mm}
\end{figure}

We demonstrate our method's ability to deal with rotational and translational degeneracies. 

The R\"umlang Construction Site dataset is collected with an ANYbotics ANYmal C robot \cite{hutter2017anymal} in R\"umlang, Switzerland. The robot traverses a large open area at a construction site, and rotates in place at locations A and B in Fig. \ref{fig:rumlang-result}. When only the ground plane in the open area is visible, translations along the ground plane and rotations about the normal of the ground plane are unconstrained. The robot is equipped with a Velodyne VLP-16 LiDAR (10 Hz). Scan-to-map ICP is only mildly affected by degeneracies when the full 360$^\circ$ field-of-view (FOV) of the LiDAR is used. Therefore, to use this dataset to study degeneracies we increase the difficulty of the estimation by reducing the horizontal FOV to 180$^\circ$. We use legged odometry to compute a prior for the LiDAR mapping module. Based on the Velodyne VLP-16's datasheet, we use 1) isotropic and uniform per-point noise models $\mathbf{\Sigma}_{p_i} = \sigma_{p}^2 \mathbf{I}$ with $\sigma_p = 1$ cm; and 2) $\sigma_i=1$ cm in the normal covariance estimation procedure. The parameter selection is based on first principles and without experimental tuning on the dataset. For \cite{zhang2016degeneracy}, we set $\lambda_{\text{min}}=24$ based on experimental tuning on the dataset.

The results are visualized in Fig. \ref{fig:rumlang-result} and summarized in table \ref{table:qualitative-results}. While the other methods exhibit degeneracy-induced drift, our method does not.

\begin{table}[b]
\vspace{-1.8em}
\centering
\caption{Qualitative evaluation of degeneracy-induced drift}
\footnotesize
\begin{tabular}{@{}lllll@{}}
Experiment         & Ours             & Zhang \cite{zhang2016degeneracy}          & Hinduja \cite{hinduja2019degeneracy}             & Lee \cite{lee2024switch}          \\ \midrule
1 R\"umlang  & $\checkmark$             & $\times$             & $\times$             & $\times$             \\ 
2 Seem\"uhle & $\checkmark$             & $\times$             & $\checkmark$             & $\times$             \\ 
3 Relyon & $\checkmark$             & $\times$             & $\times$             & $\times$             \\ 
4 Fyllingsdalen & $\checkmark$             & $\times$             & $\checkmark$             & $\times$             \\
\end{tabular}\label{table:qualitative-results}
\\
\smallskip
\hspace{2em}\raggedright
$\times$/$\checkmark$ denote presence/absence of degeneracy-induced drift.
\end{table}

\subsection{Experiment 2 - Seem\"uhle Mine}

\begin{figure}[t]
    \vspace{0.5mm}
    \centering
    \includegraphics[width=1\linewidth]{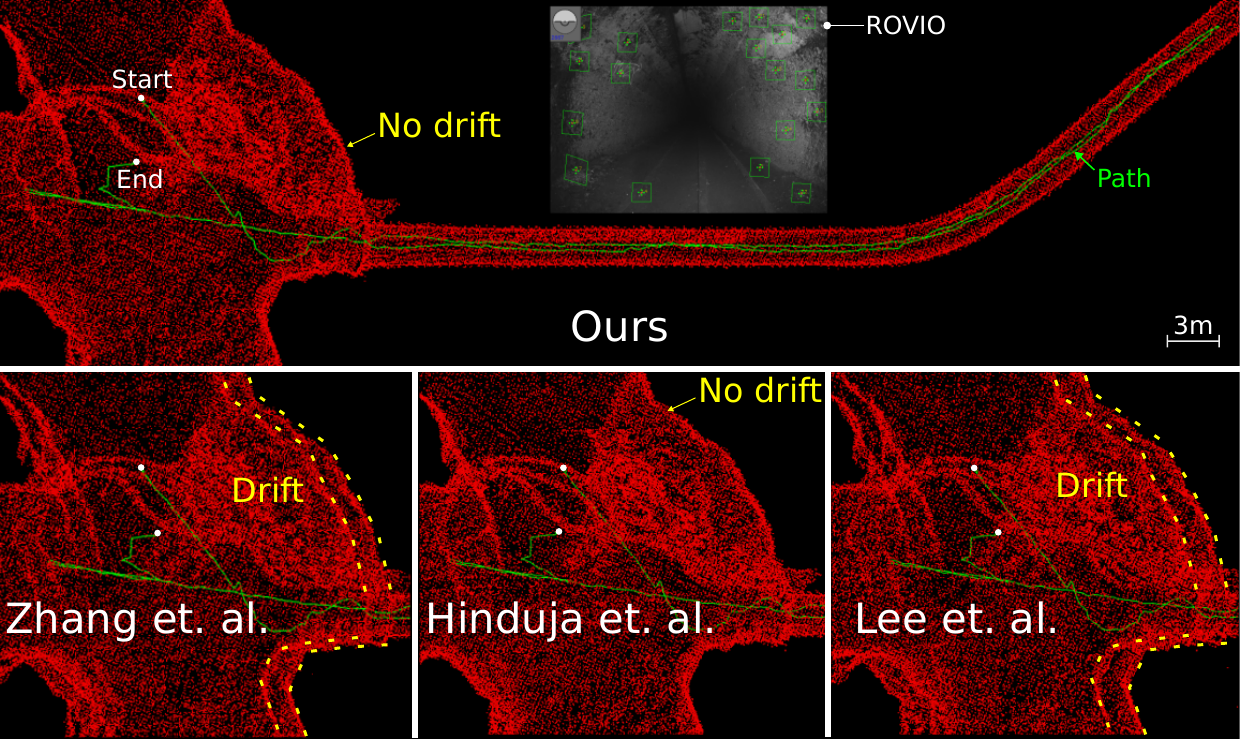}
    \caption{Partial maps of all methods from experiment 2 (Seem\"uhle), using the using the full $360^\circ$ field-of-view of the Velodyne VLP-16 LiDAR. The repeated walls in the maps of \cite{zhang2016degeneracy, lee2024switch} stem from degeneracy-induced drift.}
    \label{fig:seemulhe-result}
    \vspace{-5mm}
\end{figure}

We demonstrate the generalizability of our approach in a translationally degenerate environment, using the same platform as in experiment 1.

The Seem\"uhle Mine dataset is collected with the ANYbotics ANYmal C robot in the abandoned Seem\"uhle underground mine in Switzerland. The robot traverses a self-similar tunnel and returns along the same path. The longitudinal translation is unconstrained in most of the tunnel section. We use the full $360^\circ$ FOV of the Velodyne VLP-16 LiDAR (10 Hz). A forward-facing monochrome 0.4 MP grayscale camera (20 Hz) and the IMU (200 Hz) on an Alphasense development kit from Sevensense Robotics is used for visual-inertial odometry with ROVIO \cite{bloesch2015robust} to provide the prior for the LiDAR mapping module. A ground truth trajectory was generated for the dataset using the procedure described in \cite{tuna2023x}. The absolute pose errors (APEs) and relative pose errors (RPEs) of the estimated trajectories relative to the ground truth are computed using \cite{grupp2017evo}. Since the LiDAR, platform and the overall registration pipeline are the same as in experiment 1, we use the same parameters in experiment 2. For \cite{zhang2016degeneracy}, we double the threshold to $\lambda_{\text{min}}=48$ account for the doubled FOV of the LiDAR.

The results are visualized in Fig. \ref{fig:seemulhe-result}. Qualitative and quantitative results are summarized in tables \ref{table:qualitative-results} and \ref{table:ape-for-seemulhe}, respectively. Our method and \cite{hinduja2019degeneracy} do not exhibit visible degeneracy-induced drift and have comparable APEs in the non-tunnel segments of the trajectory (before and after the tunnel). The higher APE of \cite{zhang2016degeneracy,lee2024switch} stems from the degeneracy-induced translational drift visible in Fig. \ref{fig:seemulhe-result}. Our method yields the lowest APE overall and generalizes between experiments 1 and 2 without changing parameters. \cite{zhang2016degeneracy} yields qualitatively comparable results to our method if it is given the advantage of per-dataset tuning, resulting in $\lambda_{ \text{min}}=150$. 

\begin{table}[b]
\setlength{\tabcolsep}{4pt}
\vspace{-1.5em}
\centering
\caption{APE and RPE errors for experiment 2 - Seem\"uhle Mine}
\footnotesize
\begin{tabular}{@{}llllll@{}}
    & Metric        & Ours             & Zhang \cite{zhang2016degeneracy}            & Hinduja \cite{hinduja2019degeneracy}             & Lee \cite{lee2024switch}            \\ \cmidrule{2-6}
    \multirow{4}{*}{\rotatebox{90}{\makecell{Full \\ trajectory}}} & APE [m]        & \getvalue{apeTrlOurs} & \getvalue{apeTrlZhang} & \getvalue{apeTrlHinduja} & \getvalue{apeTrlSwitch} \\
    & APE [$^\circ$]  & {\getvalue{apeRotOurs}} & \getvalue{apeRotZhang} & \getvalue{apeRotHinduja} & \getvalue{apeRotSwitch} \\
    & RPE [m]        & \getvalue{rpeTrlOurs} & \getvalue{rpeTrlZhang} & \getvalue{rpeTrlHinduja} & \getvalue{rpeTrlSwitch} \\
    & RPE [$^\circ$]  & \getvalue{rpeRotOurs} & \getvalue{rpeRotZhang} & \getvalue{rpeRotHinduja} & \getvalue{rpeRotSwitch} \\ \cmidrule{2-6}
    \multirow{4}{*}{\rotatebox{90}{\makecell{Non-tunnel \\ segments}}} & APE [m]        & \getvalue{apeTrlOurs2} & \getvalue{apeTrlZhang2} & \getvalue{apeTrlHinduja2} & \getvalue{apeTrlSwitch2} \\
    & APE [$^\circ$]  & \getvalue{apeRotOurs2} & \getvalue{apeRotZhang2} & \getvalue{apeRotHinduja2} & \getvalue{apeRotSwitch2} \\
    & RPE [m]        & \getvalue{rpeTrlOurs2} & \getvalue{rpeTrlZhang2} & \getvalue{rpeTrlHinduja2} & \getvalue{rpeTrlSwitch2} \\
    & RPE [$^\circ$]  & \getvalue{rpeRotOurs2} & \getvalue{rpeRotZhang2} & \getvalue{rpeRotHinduja2} & \getvalue{rpeRotSwitch2} \\ \cmidrule{2-6}
\end{tabular}\label{table:ape-for-seemulhe}
\\ 
\smallskip \raggedright \hspace{0.2em} Mean (standard deviation) APE/RPE for translation [m] and rotation $[^\circ]$.\\ 
\hspace{0.2em} \hlf{Bold indicates the best results.}
\end{table}

\subsection{Experiment 3 - RelyOn Nutec Trondheim}

\begin{figure*}[t]
    \vspace{0.5mm}
    \centering
    \includegraphics[width=1\linewidth]{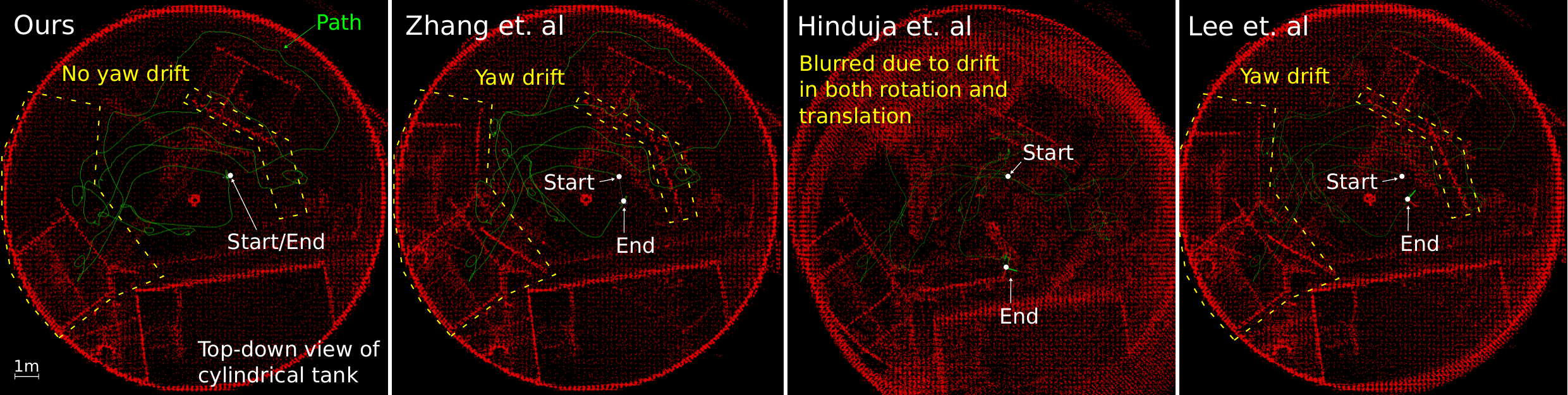}
    \caption{Partial maps of all methods from experiment 2 (RelyOn). The field-of-view of the Ouster OS0-64 LiDAR is reduced to $180^\circ$ to study degeneracies. \cite{zhang2016degeneracy,hinduja2019degeneracy,lee2024switch} exhibit degeneracy-induced drift, materializing as repeated structures in the marked areas.}
    \label{fig:relyon-result}
    \vspace{-5mm}
\end{figure*}

We further show our method's ability to deal with coupled roto-translational degeneracies and demonstrate the use of our method in factor graph-based state estimation. 

The RelyOn Nutec dataset is collected with an aerial robot in a cylindrical tank of radius $8$ m and height $16$ m at RelyOn Nutec's facilities in Trondheim. Rotations about the principal cylinder axis corresponding to height are unconstrained when only the cylinder walls are visible. In the LiDAR frame, this corresponds to the set of roto-translations such that the distance to the cylinder wall and the distance to the cylinder floor are constant. The robot is equipped with an Ouster OS0-64 RevD LiDAR (10 Hz). As in experiment 1, we reduce the FOV of the LiDAR from 360$^\circ$ to 180$^\circ$ to study degeneracies. We also use the IMU of the PixRacer Pro autopilot (100 Hz). To the benefit of the community, the dataset is released at \href{https://github.com/ntnu-arl/relyon_dataset}{github.com/ntnu-arl/relyon\_dataset}.

To integrate the LiDAR, IMU and external odometry, we make an additional extension to the LOAM mapping module by adding a LiDAR-driven factor-graph-based fixed-lag smoother using the Georgia Tech Smoothing and Mapping Library \cite{dellaert2012factor}. Three factors are added for each LiDAR frame: 1) IMU measurements are added using IMU-preintegration \cite{forster2016manifold}; 2) the estimate from scan-to-map ICP is added as unary pose prior factor; and 3) the odometry estimate is added as a binary between-factor on the current and previous pose. IMU measurements are additionally used to compute a prior for ICP. The factor-graph estimate is computed directly after scan-to-map ICP, and each scan is registered in the map by using the resulting pose estimate. The smoother lag is $3$ s.

For external odometry, we use a robot trajectory $\{ \bar{\mathbf{T}}_{L_f}^W \}_{f=1}^{N_f}$ estimated using the full FOV of the LiDAR. The raw odometry estimate between LiDAR frames $L_f$ and $L_{f-1}$ is $\Bar{\mathbf{T}}_{L_f}^{L_{f-1}} = (\bar{\mathbf{T}}_{L_{f-1}}^W)^{-1} \bar{\mathbf{T}}_{L_f}^W$. We apply on-manifold Gaussian white noise to ensure that the odometry is insufficient for successful estimation, and use the perturbed estimate $\Tilde{\mathbf{T}}_{L_f}^{L_{f-1}} = \Bar{\mathbf{T}}_{L_f}^{L_{f-1}} \text{Exp}(\theta_f)$ with $\theta_f \sim \mathcal{N}(0, \mathbf{\Sigma}_\theta)$ and $\mathbf{\Sigma}_\theta^{1/2} = \text{BlockDiagonal}(0.02 \text{ rad } \mathbf{I}_{3 \times 3},0.01 \text{ m } \mathbf{I}_{3 \times 3})$.

Based on the LiDAR's datasheet, we set the parameters of our method as follows: 1) $\mathbf{\Sigma}_{p_i} = \sigma_{p}^2 \mathbf{I}$ with $\sigma_p = 1.5$ cm; 2) $\sigma_i=1.5$ cm; 3) In the factor-graph optimization, we use the information estimate \eqref{eq:information-estimate-for-perturbation} with $\sigma_r=1.5$ cm. The tuning approach of \cite{zhang2016degeneracy} fails due to poor separation of the distribution of the minimum eigenvalue of the Hessian between degenerate and non-degenerate sections. Instead, we use a grid search and find that $\lambda_{\text{min}}=0$ yields the best results. I.e. the results are better \emph{without} this form of degeneracy handling.

The results are visualized in Fig. \ref{fig:relyon-result} and summarized in table \ref{table:qualitative-results}. Our method does not exhibit degeneracy-induced drift. All baseline methods drift in rotation. \cite{hinduja2019degeneracy},\cite{lee2024switch} additionally drift in translation, and have translational end position errors in height of $5$ m and $3$ m, respectively. The results of \cite{hinduja2019degeneracy} are particularly noisy.

\subsection{Experiment 4 - Fyllingsdalen Bicycle Tunnel}
\begin{figure*}[t]
    \vspace{0.5mm}
    \centering
    \includegraphics[width=1\linewidth]{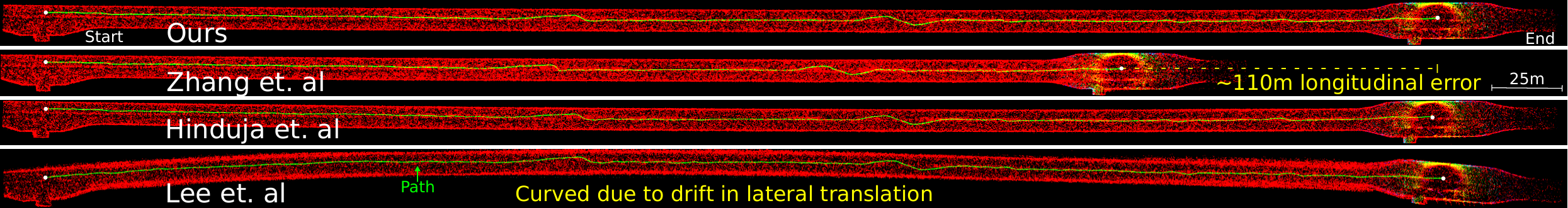}
    \vspace{-3ex}
    \caption{
    Partial maps of all methods from experiment 4 (Fyllingsdalen Bicycle Tunnel), using the full $360^\circ$ field-of-view of the Ouster OS0-128 LiDAR. \cite{zhang2016degeneracy} (second row) and \cite{lee2024switch} (fourth row) drift in longitudinal and lateral translation, respectively. Ours and \cite{hinduja2019degeneracy} (third row) yield comparable results, correctly estimating the tunnel section length to $\sim 500$ m.}
    \label{fig:fyllingsdalen-result}
    \vspace{-6mm}
\end{figure*}
We demonstrate the generalizability of our approach by validating it in a large-scale environment with translational degeneracies, using a similar platform as in experiment 3.

In the publicly available Fyllingsdalen Bicycle Tunnel dataset \cite{nissov2024degradation}, an aerial drone is flown through a 500 m straight tunnel section. The structural similarity of the tunnel renders the longitudinal translation unconstrained. The drone is equipped with an Ouster OS0-128 RevD LiDAR (10 Hz), a Texas Instruments IWR6843AOP-EVM radar (10 Hz) and a VectorNav VN100 IMU (200 Hz). 

We use the same factor-graph approach as in experiment 3, but remove the odometry factors in favor of the radar velocity factor from \cite{nissov2024degradation}. Radar velocities are estimated using \cite{doer2020ekfReve}. To avoid drift in height, we additionally use ground plane estimation to add priors on roll, pitch and height in the ICP. According to the LiDAR's datasheet, the noise characteristics of are the same as in experiment 3. Consequently, we leave the parameters of our method unchanged across the two experiments and expect the method to generalize. For \cite{zhang2016degeneracy}, the lack of useful results in experiment 3 necessitates dataset-specific tuning. Experimental tuning yields $\lambda_{\text{min}}=123$.

The results are visualized in Fig. \ref{fig:fyllingsdalen-result} and summarized in table \ref{table:qualitative-results}. With the proposed tuning, \cite{zhang2016degeneracy} underestimates the longitudinal translation by approximately $110$ m. \cite{lee2024switch} labels both lateral and longitudinal translational as degenerate in the whole experiment, and therefore exhibits significant lateral drift due to erroneous degeneracy predictions. Our method and \cite{hinduja2019degeneracy} yield comparable performance, correctly estimating the tunnel section length to $\sim500$ m. Our method generalizes between experiments 3 and 4, without changing parameters.

\subsection{Runtime}
To validate our method's ability to operate in real-time, we measure the runtime of the adapted LOAM scan-matching step in experiment 2. The timings include the k-d tree construction for the local submap before the ICP loop and per-iteration correspondence search and plane estimation. The median runtime was $17$ ms on a laptop with an Intel i7-12800H CPU. The added overhead from our method is limited as only $5.4 \%$ of the runtime is directly attributable to degeneracy detection. On the constrained Khadas VIM4 with a combined 2.2GHz quad-core Cortex-A73 CPU/2.0GHz quad-core Cortex A-53 CPU using one A73 core, the median runtime was $54$ ms.

\section{Conclusions}\label{sec:conclusion}

Detecting and handling degeneracies is critical for reliable LiDAR-based localization and mapping in challenging environments. This letter presented a new probabilistic method for detecting degeneracies in point-to-plane error minimization problems and a probabilistic degeneracy-aware iterative closest point algorithm. The proposed approach outperformed state-of-the-art methods at handling the effect of degeneracies in four real-world experiments in challenging environments. Owing to its first-principles formulation, the approach offers easier parameter selection and improved generalizability compared to the baseline methods.

\section{Acknowledgement}
We thank the Robotic Systems Lab of ETH Zurich for sharing the data from the ANYmal C robot. 

\bibliographystyle{IEEEtran}
\bibliography{IEEEabrv,manual}

\end{document}